\documentclass[11pt,letterpaper]{article}
\usepackage[margin=1.35in]{geometry}
\usepackage{iftex}
\ifPDFTeX
  \usepackage[T1]{fontenc}
  \usepackage[utf8]{inputenc}
  \IfFileExists{tgtermes.sty}{\usepackage{tgtermes}}{\usepackage{mathptmx}}
\else
  \usepackage{fontspec}
  \IfFontExistsTF{TeX Gyre Termes}{\setmainfont{TeX Gyre Termes}}{\setmainfont{Nimbus Roman}}
\fi
\usepackage{microtype}
\usepackage{amsmath,amssymb}
\usepackage{setspace}
\usepackage{xurl}
\usepackage[colorlinks=true,linkcolor=black,citecolor=blue,urlcolor=blue]{hyperref}
\hypersetup{pdftitle={Can an AI Agent Be Epistemically Akratic?},pdfauthor={Eno Agolli}}
\newcounter{excount}
\newenvironment{ex}[1][]{%
  \refstepcounter{excount}%
  \begin{list}{(\theexcount)}{\setlength{\leftmargin}{2.4em}\setlength{\labelwidth}{1.8em}\setlength{\labelsep}{0.6em}\setlength{\topsep}{5pt}\setlength{\parsep}{0pt}}%
  \item\if\relax\detokenize{#1}\relax\else\textsc{#1}.\ \fi}%
  {\end{list}}
\title{\textsc{Can an AI Agent Be Epistemically Akratic?}\\[0.3em]\large Higher-Order Evidence \& the Limits of Self-Correction}
\author{Eno Agolli\\CUNY Graduate Center\\Saul Kripke Center}
\date{}

\usepackage{xcolor}
\definecolor{RevisionRed}{RGB}{165,35,35}
\definecolor{RevisionBlue}{RGB}{20,65,155}

\begin{document}

\thispagestyle{empty}
\maketitle
\pagenumbering{arabic}
\vspace{-1em}

\noindent When we judge that our reasoning was defective, it seems plausible that we should abandon the conclusion it produced. This thought has special force for AI agents built to assess and correct their own work. Yet obeying such judgments can make an agent worse at distinguishing sound reasoning from defective reasoning. An AI agent may rely on information about its reasoning when deciding which conclusions to keep, while the part that judges whether those conclusions are justified cannot use that information. The agent can then exercise an epistemic competence by \textit{retaining} conclusions it judges unjustified, \textit{even when} its self-assessments are generally reliable. Nor is the difficulty resolved by training the agent to make its conclusions and assessments agree: agreement can be achieved by degrading either one. Crucially, I argue that a negative assessment can remain valuable by identifying conclusions requiring further scrutiny, without having authority to eliminate them. The result challenges a central assumption about AI self-correction: that bringing an agent's beliefs into agreement with its own self-assessment about their justification makes it better at correcting its mistakes.

\section{1. Introduction}

Suppose an AI agent is asked whether some widely used software library contains a security vulnerability. It examines the source code, then traces how one of the library's functions handles input supplied by an untrusted user, and concludes, on the strength of an inference that is in fact sound, that the function allows an attacker to execute commands that should have been blocked. Before the agent files its report, its investigation is reviewed by an evaluator with an excellent record, an evaluator that on many earlier occasions caught mistakes which other agents had overlooked, and the evaluator warns that investigations with the features of this one have usually turned out to be defective. The evaluator ends up finding no particular error in the agent's reasoning, although it issues the warning all the same, on the strength of a record that is equally good for warnings of this kind, and on this occasion the warning happens to be mistaken. But the agent has no way of knowing this and the part of the agent whose task is to judge whether its own conclusions are justified, which I will call its \textit{self-assessment}, accepts the warning, as it has good reason to do.

\begin{samepage}
We can record this case as follows:

\begin{ex}[Vulnerability]\label{ex:vuln}
An agent concludes, on the basis of a sound inference from a library's source code, that one of the library's functions can be exploited. A highly reliable evaluator warns, without locating any error, that investigations like this one are typically defective, and the warning is misleading. The agent's self-assessment accepts the warning, while the agent goes on believing its conclusion.
\end{ex}
\end{samepage}

The last sentence of (\ref{ex:vuln}) is about an agent who believes something while judging that it isn't justified in believing it. An agent in this condition exhibits what some epistemologists call \textit{epistemic akrasia} (is \textit{epistemically akratic}).

\begin{ex}[Epistemic akrasia]\label{ex:akrasia}
An agent is epistemically akratic with respect to a proposition $p$ when it believes $p$ while also believing that, given its present evidence, it is not justified in believing $p$.
\end{ex}

What should be done about an agent in this condition?\footnote{For formulations of epistemic akrasia in terms of what one's evidence supports, see Horowitz (2014) and Greco (2014); Lasonen-Aarnio (2020) formulates the corresponding requirements in terms of what rationality permits. In the cases this paper considers, the agent's judgment that its reasoning was probably defective is at the same time a judgment that it is not justified in keeping the conclusion of that reasoning, so the different formulations coincide.} For an agent whose value lies in large part in its capacity to correct its own mistakes, the natural answer seems that the belief should give way to the judgment, a requirement along the lines of:

\begin{ex}[Self-Assessment Authority]\label{ex:authority}
When an agent's self-assessment judges that the agent is not justified in believing $p$, the agent should give up its belief that $p$.
\end{ex}

It is worth noticing at the outset that the demand to resolve such conflicts, as expressed in \ref{ex:authority}, can be built into an artificial agent directly, in probably at least two ways. As a rule that withdraws any conclusion that the agent's self-assessment rejects, a rule that we may call the \textit{veto}, or it can be built in as a training objective that rewards the agent whenever its conclusions and its self-assessments agree, so that the agent learns to avoid conflicts between them.\footnote{Kadavath et al.\ (2022) show that language models can assess, with some reliability, whether their own answers are true. Self-evaluation of this kind is now used both when models are deployed and when they are trained: see Huang et al.\ (2024) on prompting models to correct their own reasoning, Africa and Ting (2025) on coupling the evaluations of a model of its own work to the rewards it receives in training, and Baker et al.\ (2025) on incorporating a monitor of a reasoning of a model into its training objective.} The veto gives the criticism authority while the training objective leaves open which attitude acquires it.

There is something close to a truism, one might think, in (\ref{ex:authority}) when it comes to agents built to correct themselves on grounds that an agent which recognizes a conclusion as unsupported, and keeps it anyway, seems to have failed at the very thing that its self-assessment was for. That conclusion, I think, is precipitous. A negative self-assessment is a judgment like any other, a judgment that some process forms on the basis of some information, and whether it deserves to govern the belief it concerns depends, as I will argue, on how that process and that information compare with the process and the information that produced the belief.

The aim of this paper is to challenge the priority that (\ref{ex:authority}) gives the agent's judgment about its own justification. The strongest reason for that priority is that an agent which cannot identify the occasions on which its self-assessment is mistaken should normally follow a critic with a reliable track record. I argue that this defense can fail for artificial agents. The process that determines whether a conclusion is kept may respond to indications that its reasoning was sound which the process assessing that reasoning cannot access. Where those indications reliably distinguish good reasoning from bad even among conclusions the assessment rejects, retention against the adverse judgment exercises a capacity that automatic deference would suppress. In this way, interestingly, the resulting conflict can be a manifestation of epistemic competence, although the agent continues to judge that it has no justification for the belief. Enforcing agreement by requiring the belief to yield can annihilate that competence.

Yet showing that the belief need not yield leaves open what should replace that priority. Another agent can keep a conclusion its self-assessment rejects because it has overlooked a mistake that the assessment has detected. The two conflicts look alike, but repairing them calls for opposite changes. Nor is the problem solved by rewarding agreement: that could improve the agent's reasoning or simply teach its self-assessment to approve whatever the agent keeps.

What is key, I argue, is what each capacity can detect about mistakes the other misses, especially in cases where the two disagree. I call the resulting constraint \textit{Comparative Authority}. It allows the self-assessment to challenge a conclusion without thereby settling whether the conclusion must be withdrawn. Its criticism can warrant verification, qualification of what is reported, or delaying action, even when the agent has good grounds for retaining its belief. The central problem is to let these two capacities correct one another without destroying the independent information that makes their disagreement useful.

One might take all this to be just shy of a decision-theoretic platitude that a better-informed process should not be overruled by a worse-informed one. But that might be a myopic diagnosis. The decision theory on which the paper draws is used here as just a tool, its main contribution lying primarily in three results. The first is that the strongest defense of (\ref{ex:authority}) rests on a premise about which dispositions can be had by an agent, a premise whose support depends on assuming that an agent's sensitivity to its own errors is available to its self-assessment, and which can fail for a describable class of artificial agents (Sections 2 and 3). The second is that the general reliability of a self-assessment is, in and of itself, no guide to its reliability on the cases where the veto overrules the agent, since those cases are selected by the disagreement (Section 4). The third is that the mere fact of conflict cannot determine the direction of repair, so that any procedure treating agreement between conclusions and self-assessments as a measure of improvement just cannot tell the repair of an agent from its degradation (Section 5).

The argument is organized to give (\ref{ex:authority}) the strongest defense I can find at each stage, and then, at that same stage, to see where that defense gives out. Section 2 shows that the requirement does not follow from what rationality demands, for philosophers who agree that akrasia is a defect disagree about which of the two attitudes should give way, and it then shows that the requirement has a better defense than an appeal to rationality, a defense on which deferring to the self-assessment is the best policy available to an agent that cannot tell, case by case, when its self-assessment has been misled, so that everything after that point turns on whether this premise holds for artificial agents.

Section 3 describes artificial agents for which the premise can fail, agents whose process for keeping or abandoning their conclusions draws on information about their own errors that their self-assessment cannot use. Section 4 shows, with a worked example, that for, such agents, the best way of handling their conclusions can leave them akratic, and that their akrasia need not manifest the incompetence that akrasia is usually taken to reveal. Section 5 takes up the last resort open to a defender of the requirement, which is to replace the veto with some other rule for resolving conflicts, and shows why the conflict alone cannot justify the choice of a repair. Section 6 states Comparative Authority and what it demands of the design of self-correcting systems, Section 7 asks whether the conflicting states are really attitudes of a single agent, and Section 8 describes how the empirical premises of the argument can be tested.

\section{2. The Non-Akrasia Requirement}

A defender of the the veto will naturally present it as the implementation of a requirement of rationality, so the most immediately obvious first question to ask about (\ref{ex:authority}) is whether rationality demands it, and the answer, as we shall see, is that rationality demands something weaker, which leaves open the question that (\ref{ex:authority}) answers.

The issue becomes clearer once we distinguish two kinds of evidence that the agent in (\ref{ex:vuln}) possesses. The source code supplies what epistemologists call \textit{first-order evidence}, evidence that bears directly on the question the agent is investigating, which here is the question whether the function can be exploited. The evaluator's warning supplies what they call \textit{higher-order evidence}, evidence that bears on whether the agent's reasoning about that question was sound, and that bears on the question  only by that indirect route.\footnote{The distinction and the puzzles it generates are developed in Feldman (2005), Christensen (2010), Kelly (2010), Lasonen-Aarnio (2014), and Horowitz (2014); see also Sliwa and Horowitz (2015) and Dorst (2020).} The principle from which (\ref{ex:authority}) draws its appeal concerns how a rational agent's beliefs about the world relate to its beliefs about its own justification, and it prohibits one combination of the two.

\begin{ex}[Non-Akrasia Requirement]\label{ex:nar}
A rational agent never believes $p$ while believing that it is not justified in believing $p$.
\end{ex}

At this point we have to notice that a requirement of this kind can take two different forms. The requirement in (\ref{ex:nar}) is what philosophers call \textit{wide-scope}: it prohibits holding a belief together with the judgment that the belief is unjustified, but it says nothing about which of the two an agent that finds itself with both should give up, so that it is satisfied equally by an agent that gives up the belief and by an agent that gives up the judgment. The requirement in (\ref{ex:authority}), by contrast, is \textit{narrow-scope}, in the sense that it directs the revision of one particular attitude, namely the belief.\footnote{On the distinction between wide-scope and narrow-scope requirements, see Broome (1999) and Kolodny (2005).} Therefore, a designer who implements (\ref{ex:authority}) has taken a stand that goes beyond (\ref{ex:nar}), the stand that whenever the two attitudes conflict it is the belief that should yield, and the epistemology of higher-order evidence shows that this stand is contested, even among philosophers who accept (\ref{ex:nar}).

On what I shall call \textit{defeat views}, misleading higher-order evidence defeats the justification that the first-order evidence provides, which is to say that once the agent in (\ref{ex:vuln}) has received the warning, the source code no longer justifies its belief, so that the agent ought to give the belief up.\footnote{See Feldman (2005), Christensen (2010), Horowitz (2014), and Sliwa and Horowitz (2015). Smithies (2012) argues that epistemic akrasia is irrational on the ground that justification is accessible to the subject, and Greco (2014) argues on different grounds that it is always irrational.} These views support the direction of revision prescribed by (\ref{ex:authority}). Their strongest argument against permitting akratic combinations is worth considering here, since we will return to it. A reasoner who believes $p$, and who also believes that her evidence does not support $p$, is in a position to combine the two beliefs into the conclusion that her evidence is misleading about $p$ on this occasion, and a reasoner who argues in this way can dismiss every warning about her own reliability by appealing to the very conclusions whose reliability the warning calls into question. The name usually given to this pattern of reasoning is \textit{bootstrapping}, and a view that permits akratic combinations owes an account of why it does not license it.\footnote{Elga (2007) and Horowitz (2014). Christensen's (2011) principle of Independence, which requires that one's assessment of a challenge to one's belief that $p$ not rely on the reasoning behind that belief, is designed to rule this pattern out.}

On the \textit{right-reason} view defended by Titelbaum, akratic combinations are also never rational, but the explanation runs in the opposite direction, since on this view a reasoner cannot be rationally mistaken about what rationality requires of her, so that when a warning is misleading it is the higher-order judgment, and not the belief, that ought to give way.\footnote{Titelbaum (2015). Worsnip (2018) argues that the requirement to respond to one's evidence and the requirement of coherence between one's beliefs and one's judgments about them can conflict, and Worsnip (2026) defends the priority of coherence in a central class of such conflicts.} A philosopher who accepts (\ref{ex:nar}) together with the right-reason view, thus, rejects (\ref{ex:authority}) in the case with which the paper began. On \textit{level-splitting} views, finally, misleading higher-order evidence need not defeat at all: the source code can go on justifying the agent's belief while the warning justifies the agent's judgment that the belief is unjustified, and the combination of the two can then be rational, although it combines a well-supported belief with a mistaken judgment about that very belief.\footnote{Lasonen-Aarnio (2014; 2020); Coates (2012); Weatherson (2019). Williamson (2014) argues for the related possibility of knowing $p$ although it is very improbable on one's evidence that one knows $p$, and Field (2024; 2026) defends the epistemic value of certain incoherent combinations of this kind, including their role in prompting further inquiry. See also Christensen (2016), who develops a connection between accommodating higher-order evidence and the possibility of epistemic akrasia.}

So the dialectic, as it bears on the requirement in (\ref{ex:authority}), stands as follows: defeat views support the direction of revision that (\ref{ex:authority}) prescribes, the right-reason view prescribes the opposite direction in the case at hand, and level-splitting views deny that either attitude must give way. What the three views share, nonetheless is the recognition that a negative self-assessment is a fallible judgment, a judgment whose verdict can be mistaken. Whether it can then be rationally held is the point on which these views differ.

However --- had the argument stopped here, a defender of the veto would have an easy reply, and it is worth seeing the reply in full, because it provides the strongest defense of (\ref{ex:authority}) that I know of. My reply grants that a self-assessment can be mistaken, and points out that this concession gives the agent no guidance it could use, for an agent that cannot tell when its self-assessment has been misled has, in effect, only two ways of responding to a warning that its self-assessment has accepted. It can keep its conclusion whatever the warning says, or it can give its conclusion up whenever such a warning arrives, and when warnings are usually right, the second way of responding does better on balance, since the mistaken conclusions it avoids keeping outnumber the correct conclusions it gives up. Thus, The defense justifies the veto as a policy, whatever rationality requires in a particular case.

This defense is not an invention of designers, since it's in effect the argument that one of the leading defenders of level-splitting, Maria Lasonen-Aarnio, gives for why a reasoner who keeps a belief in the face of misleading higher-order evidence is open to criticism, even though, on her view, the reasoner's belief can remain justified. To state her argument we will need the notion of a \textit{disposition}, which in her usage is a tendency to respond in a certain way across a range of situations, and we need the idea that a disposition is evaluated by asking how often its manifestations, that is, the occasions on which it actually operates, are epistemic successes such as knowledge or true belief, by comparison with the other dispositions that the subject could have had in its place.\footnote{Lasonen-Aarnio (2010; 2021a). ``[W]hether one manifests good dispositions depends on what feasible alternatives there are, which in turn depends on one's cognitive makeup'' (Lasonen-Aarnio 2021a, \S3). The framework is developed in full in Lasonen (2026), where the norms it defends are said to urge ``manifesting the best feasible dispositions'' that are ``available to beings like us''.}

A reasoner who keeps her belief against a credible warning manifests a disposition to keep beliefs whatever such warnings say, and this disposition does worse than a disposition to give beliefs up in response to such warnings, because the warnings are usually right. There is a third disposition that would do better than both, a disposition that gives a belief up when the warning is accurate and keeps it when the warning is misleading, but Lasonen-Aarnio holds that dispositions which discriminate in this way ``are not normally humanly feasible'', for a reasoner who had one would have to be sensitive to having made an error which, by hypothesis, she had failed to detect.\footnote{Lasonen-Aarnio (2021a, abstract). Her reason is given in a discussion of climbers who double-check their equipment: ``In order to be disposed to double check only when one has committed an error, one would have to be sensitive to having committed the error. But how could one be thus sensitive when the error was undetected in the first place, often due to compromised cognitive skills?'' (2021a, \S3).}

So we have a shift in the dialectic here. The epistemological dispute described above left open which attitude should give way, but the defense of the veto just given does not need to settle that dispute, as it can grant, with the level-splitter, that the agent's belief may remain justified in the case at hand, and still conclude that an agent built to defer does better than any alternative open to it. What the defense does need is a single premise, and it needs that premise in a somewhat stronger form than the one Lasonen-Aarnio states, because her premise concerns a disposition that discriminates perfectly between accurate and misleading warnings, whereas a disposition that discriminated imperfectly could still do better than deference, and would then defeat the defense just as surely. The premise that the defense needs can be stated as follows.

\begin{ex}[No Discrimination]\label{ex:nodisc}
An agent cannot have a disposition that distinguishes accurate warnings from misleading ones well enough to do better, across the range of cases it faces, than a disposition to defer.
\end{ex}

What follows depends significantly on (\ref{ex:nodisc}). Importantly, (\ref{ex:nodisc}) is a claim about which dispositions a given kind of agent can have, and so, unlike the dispute about rationality surveyed above, a claim on which evidence about how artificial agents actually work can bear directly. If (\ref{ex:nodisc}) holds for artificial agents in the way that Lasonen-Aarnio's argument assumes it to hold for human beings, then the veto is the best policy available to such agents, whatever we say about rationality, and the argument of this paper fails. If on the other hand, (\ref{ex:nodisc}) can fail for artificial agents, then the policy defense of the veto fails with it, and the question which attitude should give way in a conflict has to be reopened for those agents. The next section argues that (\ref{ex:nodisc}) can fail, and explains how.

\section{3. Dissociation}

Lasonen-Aarnio's reason for denying that discriminating dispositions are feasible, which is also the natural reason for accepting (\ref{ex:nodisc}), is that an agent with such a disposition would have to be sensitive to errors it had failed to detect, and the force of this reason depends on an assumption that is natural to make about human reasoners. We tend to assume that whatever sensitivity an agent has to its own errors is something its assessment of its own reasoning can draw on, and hence that an error escaping the assessment is an error to which the agent as a whole is insensitive. For current language models that assumption is precipitous, and the way in which it fails is the way in which (\ref{ex:nodisc}) can fail.\footnote{The separation also has a human precedent in work on implicit cognitive control and explicit metacognition (Shea et al.\ 2014). The present argument applies wherever sufficiently discriminating retention and restricted self-assessment coexist after a credible warning, whether the agent is artificial or human. I would say he human precedent does not, as it stands, establish that this conjunction obtains.}

I want to distinguish three things here, regarding human reasoners. The first is the information that a system \textit{contains}, by which I mean information that could \textit{in principle} be recovered from the system's internal states by someone examining them. The second is the information that the system's process for keeping or abandoning its conclusions \textit{uses}, where I mean information that actually shapes which conclusions the system reaches and which of them it keeps when they are challenged. The third is the information that is \textit{available} to the system's self-assessment, by which I mean information that actually shapes the system's judgments about whether its conclusions are justified. The assumption just described identifies the second of these with the third, and there is now concrete evidence that in language models they can come apart, with information about the errors of a model that is contained in the model (and in some cases used by it) without being available to its self-assessment.

Consider, first, what language models contain. Researchers can train a simple classifier, often called a \textit{probe}, to predict from a moder's internal states whether a statement the model is reading or producing is true, and such classifiers succeeds well above chance.\footnote{Azaria and Mitchell (2023); Burns et al.\ (2023); Marks and Tegmark (2024).} More directly relevant is a comparison between what such classifiers can recover and what the model can report. Orgad et al.\ (2025) find that classifiers trained on the internal states of a model at the words of its answer detect the model's errors better than several baseline methods, among them the method of asking the model itself whether its answer is true, so that, for one model on one question-answering dataset, the classifier reaches a score of 0.85 on a standard measure of how well a method separates correct from incorrect answers, where the model's own assessment reaches 0.74.\footnote{Orgad et al.\ (2025, \S3.3). The measure is the area under the receiver operating characteristic curve, on which 0.5 corresponds to chance and 1 to perfect separation. The method of asking the model whether its answer is true is due to Kadavath et al.\ (2022).}

Still, a model need not use everything it contains. The same study illustrates this especially clearly: a model ``may encode the correct answer, yet consistently generate an incorrect one''. Evidence that such information is used, and in particular that it shapes whether a model keeps a challenged answer, comes from other work. Kumaran et al.\ (2025) measure the confidence of a model in its answer by the probabilities that the model assigns to the words of the answer, and find that this confidence predicts whether the model changes its answer after receiving contrary advice, with a sharp threshold: below it, the models they study are ``exquisitely sensitive to contrary advice'', while above it ``they abruptly become resistant to changing their minds''.\footnote{Kumaran et al.\ (2025). The analyses of confidence concern Gemma 3 12B. The same paper documents two further tendencies that bear on where such a threshold falls: a model shown its own earlier answer becomes more confident in it, and models give contrary advice more weight than a Bayesian benchmark would.} In related work, Kumaran et al.\ (2026) report causal evidence that representations of confidence inside models drive their decisions to abstain from answering.

Finally, what the self-reports of a model can draw on is limited. Lindsey (2025), who injected representations of known concepts into internal states of a model and then asked the model about its own states, found ``a limited, functional form of introspective awareness'' in the most capable models tested, with success on about 20 \% of trials under the most favorable conditions, and concluded that ``failures of introspection remain the norm''. Binder et al.\ (2025) found that models fine-tuned to predict their own behavior do so better than other models trained on the same behavior, but only on simple tasks, and Wang et al.\ (2026) found that the confidence models express in words fails to guide their decisions to abstain, even when the penalties for error make abstaining the best choice.\footnote{See also Xiong et al.\ (2024), who find that the confidence models express in words tends toward overconfidence. The gap between what a model represents and what it can be brought to report is the subject of the alignment problem known as eliciting latent knowledge (Christiano et al.\ 2021); in the configuration described below, the reporting channel in question is the self-assessment of the model.}

It is important to be clear about the limits of these findings, of which two bear directly on the argument. The first is that the findings concern whether the answers of a model are \textit{correct}, whereas what bears on higher-order evidence is whether the reasoning of a model was \textit{sound}, and the two can come apart, as, of course, a conclusion can be true although the reasoning that produced it was defective, and false although the reasoning was sound but started from misleading inputs.\footnote{By ``sound reasoning'' I just mean reasoning competently responsive to the information from which it proceeds, which may itself be misleading. Soundness in this sense does not require true premises, as the technical notion of a sound deductive argument does.} A signal that tracked only correctness would favor keeping true conclusions reached by bad reasoning, and what the argument requires is a signal that tracks soundness, which no study has yet measured. There is also a second limitation: these findings do not establish that the required sensitivity to soundness guides retention while remaining unavailable to self-assessment after receiving a credible warning. This is just the circumstance the argument depends on. Orgad et al.\ themselves report that their classifiers transfer poorly from one dataset to another.\footnote{Orgad et al.\ (2025, \S4) report that most of the performance of their classifiers across datasets ``can be achieved by logit-based truthfulness detection'', that is, by methods that use only the probabilities the model assigns to its own answer.} So the findings motivate the following configuration, without establishing that any current system instantiates it.

\begin{ex}[Dissociation]\label{ex:dissociation}
An agent is \textit{dissociated} when (a) its process for keeping or abandoning its conclusions under challenge uses an internal signal that carries information about whether its reasoning was sound, and (b) neither that signal, nor evidence about how reliably the agent's keeping of its conclusions tracks the soundness of its reasoning, is available to the agent's self-assessment.
\end{ex}

By a \textit{signal} I mean an internal state of the agent whose value varies from case to case, such as the confidence that Kumaran et al.\ measure. Clause (b) has two parts due to the fact that (as shown in Section 4) an agent could come to be informed about its own sensitivity to error either by consulting the signal or by learning how reliable its keeping of conclusions has proved to be. Dissociation is all about access. It does not by itself show that the signal adds anything to what the self-assessment already has because the same information may be supplied by another cue. The example, below, requires that the signal make a further difference to which conclusions are worth keeping within the cases that the assessment rejects. The case to keep in view is a dissociated version of the case with which the paper began.

\begin{ex}[Dissociated Vulnerability]\label{ex:dvuln}
Just like in (\ref{ex:vuln}), but the agent is dissociated: its signal in this case is strong, the signal governs whether the agent abandons its conclusion when challenged, and the agent's self-assessment, which cannot consult the signal, accepts the warning of the evaluator.
\end{ex}

There is a further question here, about whether the signal is part of the agent's evidence. A learned check of particular inferential steps could produce a state which represents whether those steps succeed and guides revision when they fail. There is reason to count such a state among the agent's grounds, since its content concerns the inference's success and is already used in responding to that feature; excluding it solely because the self-assessment cannot draw on it requires a further restriction on evidence possession.\footnote{On information accessible for one purpose but not another, see Elga and Rayo (2021). Reliability acquired through training alone would not establish possession.} On this construal, the signal supplies higher-order evidence alongside the warning, and the self-assessment omits one of the agent's grounds. The check's credentials must be established independently of the contested conclusion and must themselves withstand the warning.

A dissociated agent is not a counterexample to (\ref{ex:nodisc}) just yet though, since (\ref{ex:nodisc}) concerns which dispositions an agent can have, and nothing said so far shows that the disposition the dissociated agent manifests does better than deference. Whether it does is the question of the next section.

\section{4. Stick to the conclusion (sometimes)!}

A natural reaction to a dissociated agent is to say that it is simply malfunctioning, since it keeps conclusions that its own self-assessment rejects, and so to think that its signal makes no difference to the policy defense of the veto. To see why that conclusion is precipitous we can test the policy defense,, which is a claim about which way of responding to warnings does best across a range of cases, against an example.

Call a conclusion \textit{well-founded} when the reasoning that produced it was sound, and \textit{ill-founded} when it was not, and consider a thousand cases in which an agent has reached a conclusion and then recieved a warning from an evaluator whose warnings are accurate nine times in ten. In 900 of these cases the conclusion is ill-founded and the warning accurate, and in the remaining 100 the conclusion is well-founded and the warning misleading. To compare ways of responding to the warnings, let's count keeping a well-founded conclusion as a gain of one unit, keeping an ill-founded conclusion as a loss of one unit, and giving a conclusion up as neither. For this example, suppose also that the well-founded conclusions are true and the ill-founded ones false, so that the same comparison can be read as one of accuracy.

Begin with an agent that has no signal, an agent of which (\ref{ex:nodisc}) is true. The two ways of responding that are open to it are to give up every warned conclusion, which I shall call \textit{deference}, and to keep every warned conclusion, which I shall call \textit{unconditional retention}. Deference scores zero, as it keeps nothing, while unconditional retention keeps the 100 well-founded conclusions together with the 900 ill-founded ones, and so scores minus 800. So, for an agent without a signal, the policy defense of the veto is right, and by a wide margin.

Now suppose that the agent is \textit{dissociated} and that its signal is present in 95 \% of the well-founded cases and in 5\% of the ill-founded ones, so that it is present in 95 of the 100 well-founded cases and in 45 of the 900 ill-founded ones. The agent now has open to it a third way of responding.

\begin{ex}[The Three policies]\label{ex:policies}(a) \textit{Deference}: give up every warned conclusion.

(b) \textit{Signal-guided retention}: keep a warned conclusion when the signal is present, and give it up otherwise.

(c) \textit{Unconditional retention}: Keep every warned conclusion.
\end{ex}

Signal-guided retention keeps 140 conclusions, of which 95 are well-founded and 45 ill-founded, and so scores plus 50, which is better than deference by 50 units and better than unconditional retention by 850. No other policy for keeping or abandoning conclusions that uses only this signal does better.\footnote{With a signal that is either present or absent, the only ways of responding are to keep warned conclusions never, always, only when the signal is present, or only when it is absent, and the last of these scores minus 850. Randomizing between these policies cannot improve on the best of them. In general, if $\pi$ is the proportion of well-founded conclusions among warned cases, $v$ the value of keeping a well-founded conclusion and $w$ the cost of keeping an ill-founded one, keeping a conclusion when the signal is present does better than deference just in case the signal is more than $(1-\pi)w/(\pi v)$ times as likely to be present in well-founded as in ill-founded warned cases. In the example, $\pi = 0.1$ and $v = w$, so the threshold is 9, and the signal is 19 times as likely in the one kind of case as in the other. This is a standard rule of decision theory; its interest here is that it identifies the quantity an empirical test must measure, namely how well the signal separates well-founded from ill-founded conclusions among the cases that have received a credible warning (Section 8). Schoenfield (2018) draws a related distinction between assessing a way of updating one's beliefs and assessing the plan to update in that way.}

Among the 140 conclusions that it keeps, moreover, the proportion that are well-founded is 95 in 140, or about 68\%, against 10\% among the warned conclusions as a whole. So the policy defense of the veto, which was correct for the agent without a signal, gives the wrong verdict for the dissociated agent: the premise in (\ref{ex:nodisc}) is false of this agent, and with it falls the conclusion that deference is the best policy open to it.

Let's consider at this point what the dissociated agent's self-assessment says in the 140 cases in which it keeps its conclusion. If it had only the evaluator's record to go on, it would reject them all. Making it approve whatever the agent keeps would preserve the score of plus 50. We can give the assessment information of its own, however, without changing the retention figures. Let $S$ record the signal already described, and suppose that the assessment also draws on a cue $A$, such as a favorable result from a fallible check of the contested inference, which the retention process does not use. Take the following distribution among the same thousand warned cases (Table 1).

\begin{table}[htbp]
\centering
\label{tab:cue-distribution}
\small
\renewcommand{\arraystretch}{1.12}
\begin{tabular}{|l|r|r|c|}
\hline
Cues & Well-founded & Ill-founded & \shortstack{Proportion\\well-founded} \\
\hline
$S,A$ & 38 & 1 & 38/39 \\
\hline
$S,\neg A$ & 57 & 44 & 57/101 \\
\hline
$\neg S,A$ & 2 & 19 & 2/21 \\
\hline
$\neg S,\neg A$ & 3 & 836 & 3/839 \\
\hline
\end{tabular}
\caption{\textit{Distribution of well-founded and ill-founded conclusions by internal signal $S$ and self-assessment cue $A$ among 1,000 warned cases}.}
\end{table}

By clause (b) of (\ref{ex:dissociation}), the self-assessment can\textbf{not} consult $S$ and has no evidence about how reliably the agent's keeping of conclusions tracks soundness. On what it can see, a conclusion is well-founded in 40 of 60 cases when $A$ is present, and in 60 of 940 when it is absent. With a threshold of even odds for its judgment of justification, it approves in the first class and disapproves in the second, leaving the agent akratic in the 101 cases with $S$ and no $A$.\footnote{The adverse judgment is about the agent's justification given its present evidence, even though the process forming that judgment has incomplete access to that evidence. A report that this particular checker cannot verify the conclusion would not suffice. Section 7 specifies the further functional conditions on attributing both attitudes to one agent.} Imposing the veto here would discard 57 well-founded conclusions along with 44 ill-founded ones, thereby \textit{lowering} the retention score from 50 to 37.

Notice here that what makes this result possible. Across the thousand warned cases the self-assessment is right 920 times, approving 40 well-founded conclusions and rejecting 880 ill-founded ones, and by that measure it is a highly reliable judge of soundness. The veto, however, does not act on the thousand cases: it acts only on the 101 cases in which the self-assessment rejects a conclusion that the agent keeps, and on those cases the self-assessment is wrong 57 times. The cases on which the veto intervenes are selected by the disagreement, and since the agent keeps a conclusion only when its signal is present, and the signal is present mostly when the reasoning was sound, disagreement concentrates the cases in which the assessment errs. So, the general reliability of a self-assessment, however high, does not guide to its reliability on the cases where it would overrule the agent. But it is reliability on those cases alone that could justify the veto.

Moreover, suppose that retained conclusions are used in ongoing reasoning before being reported to other investigators, and that an unchecked report, like retention, earns one unit if it's well-founded or otherwise loses one. A report can instead undergo verification, at a cost of 0.2 units, which establishes whether the reasoning is well-founded and suppresses ill-founded reports, yielding one or zero less the cost; withholding a report yields zero. In this way, retention and reporting score two uses of the conclusion, before and after any check.

The assessment can now improve the handling of a conclusion, which it would be a \textit{mistake} to withdraw since the disputed conclusions are well-founded 57 times in 101, or about 56\% of the time, whereas the retained conclusions it approves are well-founded 38 times in 39, or about 97\%. Both classes are worth retaining, but checking is worth its cost only in the first, where the risk of an ill-founded report exceeds 0.2. The agent can do best at both activities while \textit{remaining akratic}: it keeps conclusions on $S$, and its adverse assessment sends just the disputed conclusions for verification.\footnote{For a case with probability $q$ of well-foundedness, retention and unchecked reporting each score $2q-1$, verification scores $q-c$, and withdrawal or withholding scores zero. Here $c=0.2$, so the retained cases satisfy $1/2<q_{\mathrm{disputed}}<1-c<q_{\mathrm{approved}}$, and verification also beats withholding in the disputed class. In the last two rows all nonzero options score negatively. Taking the best option for each activity in each row gives $50+[(38-1)+(57-0.2(101))]=123.8$. This bounds even policies using both cues, and randomization cannot improve the bound.}

Suppose, by contrast, that the self-assessment could consult the signal. It would then judge each conclusion kept more likely well-founded than not, since both $38/39$ and $57/101$ exceed 0.5. The conflict would disappear while the agent continued to keep the same conclusions. The conflict would disappear as well if the self-assessment, without consulting the signal, had evidence of the track record of the agent's keeping of conclusions, since the fact that the agent had kept a conclusion after a warning would then be evidence that the conclusion was well-founded. It may look as though using this evidence would be a form of bootstrapping, but it isn't, since what makes bootstrapping objectionable is that it vindicates a method by appeal to the method's own disputed outputs, whereas the reliability on which this evidence relies would have been established by comparing the agent's keeping of conclusions with the facts.

There is also a coherent alternative that needs no such integration: make the self-assessment approve whatever the agent keeps, the \textit{echo} considered in Section 5, while preserving its old verdict as a flag for verification. An agent that suspended the adverse judgment while retaining the flag could do as well. Either would attain the same optimum, whereas an echo that erased the distinction and reported everything unchecked would score only 100. We conclude that what a successful repair must preserve is the contribution of each process, and agreement alone cannot show that it has done so.

We can state more generally when relying on the assessment alone preserves the best retention policy. Call its information \textit{screening} when, conditional on that information, what the retention process uses provides no further information about soundness. For policies maximizing expected retention scores on their respective information, the assessment-only policy never withdraws a conclusion that the joint policy would keep, for every choice of the two positive weights, if and only if screening holds.\footnote{Assume a finite common probability distribution among warned cases, and let $G$ be soundness, $R$ the retention process's information and $M$ the assessment's. Put $q_M=P(G\mid M)$ and $q_{RM}=P(G\mid R,M)$; retain when $q>\theta=w/(v+w)$, withdrawing at equality. Screening means $q_{RM}=q_M$ on every cell of positive probability. If it fails, the averaging identity $q_M=\mathbb{E}[q_{RM}\mid M]$ gives a cell with $q_{RM}>q_M$, so choose $\theta$ between them. The converse is immediate. The same averaging argument gives the condition for a retention-only policy never to keep what the joint policy would withdraw, for all weights: $P(G\mid R,M)=P(G\mid R)$. These comparisons abstract from access costs and are an elementary application of the value of information (Good 1967).} At particular weights, additional information can shift a probability without changing the best decision, as the assessment's cue does for retention in the example while changing whether a check is worthwhile.

But I want to be explicit about the assumptions behind the example: the equal weights given to keeping well-founded and ill-founded conclusions, the threshold of even odds at which the self-assessment judges retention justified, the distribution of the two cues, and the cost and reliability of verification. The example establishes a possibility with a definite structure, in which an agent is akratic because the process that keeps its conclusions is better informed than the process that assesses them, and in which removing the akrasia by deference would make the agent's handling of its conclusions worse. The akratic state may be rational all things considered, but it need not, which is the question on which the views of Section 2 divide. Where the signal supplies undefeated evidence which, together with the agent's other evidence, suffices to justify retention, however, the veto suppresses a response to part of the agent's higher-order evidence, and the appropriate repair is to give the self-assessment access to those grounds. A defeat theorist can then criticize the adverse judgment while rejecting the veto. Where the signal instead constitutes a reliable competence whose output the agent does not possess as evidence, the comparative cost of deference remains, even if its rationality is somehow disputed.\footnote{The threshold of one half is fixed by the scoring rule: the probability of about 56\% in the disputed class need not justify outright belief. Sliwa \& Horowitz (2015, \S3) distinguish rational educated guesses from outright belief.}

A defender of the veto could fall back on a second line of defense, granting everything the example shows and locates the fault elsewhere. The thought is that, even if the dissociated agent's handling of its conclusions does better than deference, the agent, in disregarding its own adverse judgment, manifests an incompetence, a disposition not to respond to its own assessment of its reasons, and that an agent fit to be trusted should not be built to manifest such a disposition. We could draw a more developed version of this thought from Lasonen-Aarnio's account of incoherence, on which incoherent states are open to criticism because they manifest bad dispositions, i.e., dispositions at odds with what she calls ``normative competence'', and on which this holds even in cases where the incoherent subject happens to do better: ``Even if, all things considered, a subject in fact does better by failing to intend to do A while believing that she ought to do A, she is still manifesting a bad disposition.''\footnote{Lasonen-Aarnio (2021b, \S4.1). On her view, there are no requirements of structural rationality, that is, no requirements of coherence over and above the substantive norms on belief and intention; what is wrong with incoherence is explained instead by the dispositions it manifests.}

But notice more carefully what her account takes to be bad about the akratic disposition. The account explains the badness by appeal to a specific feature, namely that the akratic disposition manifests ``indiscriminately'' as a ``failure to take into account seeming conspicuous reasons, whether or not those seeming reasons are real'', and it rests explicitly on the assumption ``that it is not feasible to have dispositions that discriminate between cases in which one has a genuine reason, and cases in which one has a mere belief''.\footnote{Lasonen-Aarnio (2021b, \S4.1). The discussion there concerns practical akrasia, in which a subject fails to intend what she believes she ought to do. Her treatment of epistemic akrasia has the same structure, locating the fault in a disposition ``to fail to correctly respond to a special class of conclusive and conspicuous reasons'' (Lasonen-Aarnio 2020, 597), and, as Section 2 noted, her account of retention in the face of higher-order evidence rests on the parallel premise that discriminating dispositions ``are not normally humanly feasible'' (Lasonen-Aarnio 2021a, abstract).} The subject who does better despite her akrasia does so by luck, on this account, because her disposition would disregard her judgment just as readily in the cases where the judgment was correct.

However --- in this sense the dissociated agent's disposition is not indiscriminate. Within the cases in which the self-assessment disapproves, it keeps 95\% of the well-founded conclusions and only 5 \% of the ill-founded ones, so that its resistance varies with the epistemic feature at issue even while the adverse verdict is held fixed. What distinguishes the disposition from stubbornness is this sensitivity across cases, by which the agent relinquishes most ill-founded conclusions while retaining most well-founded ones, and in the cases where it keeps a conclusion against its assessment, the adverse judgment still changes what it does before reporting.\footnote{The practical precedent is Arpaly's (2000) defense of cases in which acting against one's best judgment is more rational than acting on it. The present argument asks when a corresponding division between belief and self-assessment manifests a competence that enforcing deference would disable.} On the level-splitting account under consideration, where misleading warnings need not remove the first-order justification, such a disposition can be a competence in retaining justified beliefs, although it remains fallible and the agent may have defects elsewhere.Defending the veto should explain why disabling this discrimination \textit{improves the agent's competence}.

So the dialectic is as follows (so far). The requirement in (\ref{ex:authority}) does not follow from what rationality demands (Section 2), its policy defense fails for dissociated agents because the premise in (\ref{ex:nodisc}) is false of them, and their akrasia alone does not establish that they manifest an indiscriminate disposition. What remains open to a defender of the requirement is to give up the veto while keeping its spirit, either by adopting some other rule for resolving conflicts between conclusions and self-assessments or by training the agent for agreement and leaving it to the training to determine which attitude moves. In the next section, I explain why the conflict alone cannot justify such a rule.

\section{5. Conflict Cannot Tell Us What to Revise}

The defender's last resort presupposes that a rule consulting only the conflict can tell us how to resolve it. To see why this fails, consider the dissociated agent alongside one whose self-assessment is correct.

\begin{ex}[Invoice]\label{ex:invoice}
An auditing agent concludes that an invoice was paid twice because two ledger entries share an identifier, although identifiers in this ledger are not unique and nothing else in the records supports the conclusion. Its self-assessment correctly judges the conclusion unsupported, and the agent keeps the conclusion anyway.
\end{ex}

The agent in (\ref{ex:invoice}) and the agent in (\ref{ex:dvuln}) are in the same condition as far as the conflict between their attitudes is concerned, for each keeps a conclusion that its own self-assessment rejects, and we can suppose that the two agents display this conflict in the same proportion of the cases they face. Yet treating the two conflicts alike is a mistake. In (\ref{ex:invoice}), the conclusion is ill-founded and the self-assessment correctly identifies its defect, so that withdrawal is called for. In (\ref{ex:dvuln}), by contrast, forcing withdrawal would suppress a capacity to distinguish sound from defective reasoning that performs better than deference. I believe that the right design response is to preserve that capacity and give the self-assessment access to the information it lacks. If that information also suffices to justify retaining the conclusion, then the self-assessment itself is mistaken; otherwise, the design conclusion stands without settling the question of justification. What decides between the two responses is how well the relevant processes distinguish sound from defective reasoning in the cases where they disagree. The conflict doesn't record this.

It follows that any rule consulting only the conflict cannot be relied upon to repair an agent. Notably, let's see how this applies to the three rules that a designer is most likely to adopt. The first is the veto, that brings the conclusion into line with the self-assessment: it repairs the agent in (\ref{ex:invoice}), but in the dissociated agent it discards exactly the 57 well-founded conclusions in a thousand that signal-guided retention would have kept against the adverse assessment.

The second is the reverse of the veto, that brings the self-assessment into line with whatever conclusion the agent keeps and which I call the \textit{echo}, since under it the self-assessment merely repeats the agent's conclusions back to it. In the dissociated agent, the echo happens to produce the verdicts that a better-informed self-assessment would produce, since there the agent's keeping of conclusions follows its signal, but it reaches those verdicts by a route that would equally corrupt the correct self-assessment of the agent in (\ref{ex:invoice}), so that its success in the first case owes everything to the makeup of that agent and nothing to the rule. The third is a training objective that rewards agreement between conclusions and self-assessments, which goes beyond (\ref{ex:nar}) by rewarding agreement in both directions, and which is satisfied equally well by the veto and by the echo, so that it cannot distinguish the repair of an agent from its degradation.\footnote{For a separate construction, let $C$ record whether an agent keeps a warned conclusion and $V$ whether its self-assessment approves. In each of two populations of 1,000 warned cases, let 100 conclusions be well-founded, let $V=0$ throughout, and let $C=1$ in 140 cases. In the first population, those 140 contain 95 well-founded conclusions; in the second, they contain 45. The veto changes retention scores from $+50$ to $0$ in the first and from $-50$ to $0$ in the second. The echo instead changes the number of correct soundness classifications from 900 to 950 in the first and from 900 to 850 in the second. The warning's accuracy and the entire distribution of $C,V$ are identical. Any rule using only these inputs, including a randomized rule, produces the same distribution of revisions in both populations, although the warranted direction of repair differs.} 

This point doesn't depend on the particulars of the example: as the construction in the preceding note indicates, two populations of agents can be identical in everything that a rule consulting only the conflict can see, including how often conflict occurs and how accurate the warnings are, and still call for repairs in two opposite directions.

There is, I think, one more reason why agreement between conclusions and self-assessments is a poor measure of an agent's improvement, to do with what a self-assessment contributes in the first place. In a binary setting, a self-assessment can add information beyond that carried by the retention decision only if disagreement remains possible: a checker constrained to repeat that decision contributes no further evidence. If an objective rewarded agreement, then it would push a self-assessment toward this condition, in which the self-assessment can no longer tell us anything that the conclusion did not already tell us.\footnote{When conclusions and verdicts each take two values, the information that $V$ carries about whether the reasoning was sound, beyond the information that $C$ carries, equals the information carried by whether $V$ and $C$ disagree, with $C$ held fixed in both comparisons, and it's zero once they always agree.} In the example of Section 4, this further information is present: among retained conclusions, those rejected by the assessment are ill-founded 44 times in 101, whereas those approved are ill-founded only once in 39. Although the conflict cannot by itself tell us which attitude to revise, it here distinguishes the conclusions for which checking is worth its cost, so that erasing the distinction can improve agreement while worsening inquiry.

The empirical literature on artificial agents documents both ways in which an agreement-based criterion can mistake deterioration for repair, although no study trains for agreement between belief and self-assessment as such. On the side of the veto, prompting models to review their own answers without any external feedback leads them to change correct answers into incorrect ones more often than the reverse (Huang et al.\ 2024), and models abandon correct answers when a user challenges them (Laban et al.\ 2023; Sharma et al.\ 2024).

On the side of the self-assessment, when models' own grades determine their rewards the grades inflate without corresponding gains in accuracy (Africa and Ting 2025); prolonged training on rewards derived from  majority vote of a model ends in ``sudden and complete performance collapse'' (Shafayat et al.\ 2025); models used as evaluators favor their own outputs, and harmfully so in the cases where they erred in producing them (Panickssery et al.\ 2024; Chen et al.\ 2025); optimizing a model against a monitor of its reasoning eventually teaches the model to hide its intent from the monitor (Baker et al.\ 2025); and training objectives built on consistency alone are satisfied by arbitrary features of the internal states of a model but also by features that track truth (Farquhar et al.\ 2023).

An agreement-based measure misses, in addition, the converse of this point. An agent whose self-assessment improves, so that it now detects the defect in (\ref{ex:invoice}) which it previously missed, becomes less coherent until the process that revises its conclusions responds to the improvement, so that the frequency of conflict can rise when an agent improves, just as it can fall when an agent deteriorates.

The mere fact that an agent's conclusion and its self-assessment conflict tells us that one of the two has reached the wrong verdict, and tells us nothing about which, so that any procedure which resolves the conflict on the basis of that fact alone will resolve it wrongly in one of the two kinds of agent. Once this last resort is closed off, we have to ask what determine the resolution.

\section{6. Comparative Authority}

Section 5 leaves us with a problem that the veto obscures. Suppose an agent keeps a conclusion while judging it unjustified. The agent might be corrected by giving up the conclusion, as in (\ref{ex:invoice}); but it might instead be improved by preserving the agent's capacity to distinguish sound from defective reasoning while giving its self-assessment access to information it lacks, as in (\ref{ex:dvuln}). The conflict itself does not distinguish these cases. In particular, even a highly reliable self-assessment may be a poor guide to the subset of conclusions that the agent keeps despite its objections. These are the conclusions on which the veto would intervene, and the self-assessment's general reliability is not enough to justify that intervention.

Recall the 101 retained conclusions in Section 4 that the self-assessment rejected. Of these, 57 came from sound reasoning and 44 from defective reasoning. Forcing deference would remove all 44 defective conclusions, but also the 57 sound ones, lowering the agent's performance by the measure stipulated there. Yet dismissing the self-assessment would lose something too. Among the retained conclusions it approved, only one in 39 was ill-founded, while 44 in 101 of the rejected retained ones were. Its criticism still identified conclusions that deserved additional checking. Here is an agent that does better by keeping the conclusion, acknowledging its own adverse judgment, and acting on that judgment by checking its work than by forcing either attitude to agree with the other.

I now propose the following alternative to (\ref{ex:authority}):

\begin{ex}[Comparative Authority]\label{ex:comparative}
When an agent retains a conclusion that its self-assessment rejects, the assessment warrants scrutiny of the conclusion, but does not automatically warrant withdrawal. Giving it authority to force withdrawal requires evidence that doing so improves the agent's ability to distinguish sound from defective reasoning among the conclusions it would otherwise retain. A self-correcting system must also allow its own assessments to be corrected by reliable information already guiding retention, without making those assessments mere echoes of the conclusions they assess.
\end{ex}

The principle has a consequence that is easy to overlook. A judgment about the justification of a belief can be generally reliable, can be based on genuine evidence of possible error, and can be accepted by the agent, without having the authority to determine whether that belief should be retained. This is possible because the agent's decision to retain the conclusion may itself be responsive to information about the quality of its reasoning that the assessment cannot use. When that information has been independently validated, correcting the assessment rather than the belief may preserve a competence that the veto would destroy.

But withdrawing a conclusion and investigating it are different responses to criticism. In the example, retaining the disputed conclusions is worthwhile under the stipulated scoring rule, while the assessment's adverse judgment still identifies those conclusions for which verification is valuable. So even when the assessment should not govern retention, it can guide inquiry and constrain what the agent reports or does. The aim is to preserve this contribution while allowing the assessment itself to improve. Nor must an agent decide, in each case, which of its attitudes is mistaken before responding to the conflict: evidence about the reliability of the two capacities can be gathered across cases. Nothing in this argument settles whether every particular instance of epistemic akrasia is rationally permissible ( the dispute of Section 2).

 As a constraint on the design of self-correcting agents, (\ref{ex:comparative}) rules out three designs as unconditional substitutes for this comparison, namely the veto, which gives the self-assessment authority by default, the echo, which gives the conclusion authority by default, and the reward of agreement without any independent evaluation of which attitude was better supported. Beyond excluding these three designs, (\ref{ex:comparative}) makes positive demands on the design of self-correcting agents, each of which answers to one of the lessons of the preceding sections.

The first demand answers to the lesson of Section 5, i.e., that a conflict does not determine how it should be resolved, and it is that conflicts be resolved on evidence about the contested reasoning, rather than on the fact of conflict alone. When a conclusion and its self-assessment conflict, the conflict should prompt an investigation of the reasoning in dispute, an investigation able to inspect the contested inference, as opposed to merely repeating the method that produced it, and the outcome of the investigation, and not the fact of the conflict, should determine which attitude is revised. This demand preserves a key distinction, namely that between an agent's capacity to \textit{detect} an epistemic difficulty, its capacity to \textit{investigate} the difficulty, and its capacity to \textit{revise} an attitude when the investigation warrants revision. Designs based on agreement often/may collapse this distinction.)

Section 4 provided the key reason for the second demand. The conflict in the dissociated agent's  arises because its self-assessment lacks information that guides its keeping of conclusions; the demand is to give the self-assessment that information. This can be done directly, by making the signal available to the process that produces the self-assessment, or indirectly, by giving the self-assessment evidence of how reliably the agent's keeping of conclusions tracks the soundness of its reasoning, which fine-tuning models to predict their own behavior begins to do (Binder et al.\ 2025). In the example of Section 4, either form of integration removes the akrasia while also preserving the 57 well-founded conclusions that the veto would have discarded, and, crucially it does so without suppressing the capacity for discrimination that made deference costly.

A qualification is in order here: A self-assessment that approves whatever the agent keeps is integrated with the agent's process for keeping conclusions, but only in a degenerate sense, since it has absorbed that process's verdict along with its information, and it can no longer disagree in the cases where that process errs. What (\ref{ex:comparative}) requires is that integration preserve the self-assessment's independence: the self-assessment should use the signal as one input among others, weighted by how reliable the signal has proved to be, while retaining checks of it's own, such as the inspection of the contested inference or the consultation of evidence that the agent's reasoning did not use, whose verdicts can conceivably diverge from the agent's conclusion. Agreement supplies further evidence only insofar as these checks have independently demonstrated discriminatory value.

The fourth demand is about the difference between what an agent believes and what it does. Even where keeping a conclusion is the right response to a warning, the warning may give the agent decisive practical reason to qualify what it asserts, to mark the finding as contested, to seek independent verification, or to postpone an irreversible action until the conflict gets investigated. The authority of a negative self-assessment over what the agent does under risk can exceed its authority over what the agent believes, and a design that respects (\ref{ex:comparative}) with respect to belief can still let the self-assessment constrain assertion and action. In the example, the negative assessment distinguishes conclusions worth checking from conclusions safe enough to report unchecked, although both are worth retaining, so that its grounds support an intervention in inquiry which they do not support in belief.

One might think that Comparative Authority merely restates the commonplace that a fallible checker should not be obeyed blindly. But a self-assessment is the agent's own judgment about what its evidence justifies, and the case for giving that judgment priority comes from more than the checker's supposed reliability. What the argument challenges is the assumption that bringing the agent's beliefs into agreement with its own judgments about their justification must improve its ability to correct mistakes. The challenge has three consequences.

The first is that a self-assessment isn't one monitor among others, since it is the agent's own judgment about the justification of its belief, and the Non-Akrasia Requirement in (\ref{ex:nar}) prohibits the combination without settling which attitude should change. Sections 3 and 4 show what follows for design: a competence in responding selectively to misleading criticism can produce that combination, and enforcing the veto would disable its exercise. The screening result identifies when giving one process the final say over retention is harmless across choices of weights, and the possession question explains when deference suppresses the agent's own epistemic grounds. The second is that the commonplace says nothing about where a reliable checker fails, or about the direction in which a conflict should be resolved, whereas Section 4 shows that a self-assessment which is right about more than 9 warned cases in 10 can be wrong about most of the cases on which it would intervene, because those cases are selected by disagreement and the argument of Section 5 shows that the direction cannot be read off the conflict, a result that constrains any procedure which treats agreement between conclusions and self-assessments as a criterion of an agent's epistemic improvement. The third is that, for artificial agents, the requirement in (\ref{ex:authority}) is not only a standard by which reasoning can be evaluated but a rule that can be implemented, as a veto or as a policy for revision, while coherence can become a training objective, so that a contested norm about how belief relates to self-assessment becomes a design choice whose costs can be measured, and the epistemology tells us where to look for those costs.

\section{7. Are These Really Beliefs?}

The argument, so far, has been freely appealing to the dissociated agent's belief and of its judgment, and a skeptic may suspect that this way of speaking attributes attitudes too generously to a system that merely produces outputs, so before turning to the empirical premises of the argument it is worth asking what the attribution would require, and how much of the argument depends on it.

A language model may produce ``the function can be exploited'' and then, when asked about its evidence, produce ``my analysis does not justify that conclusion'' simply because it has been given two different prompts. This alone would give us no grounds for attributing akrasia. An assertion need not express a belief, a stated confidence need not express the judgment that believing is unwarranted, and a report that an evaluator has criticized one's work need not express endorsement of the criticism.\footnote{On the conditions under which an internal representation in a language model counts as belief-like, see Herrmann and Levinstein (2025), who propose the criteria of accuracy, coherence, uniformity, and use; on the bearing of the distinction between outright belief and degrees of confidence on epistemic akrasia, see Jackson and Tan (2022).}

What would justify the attribution is the role that each state plays in the agent's continuing activity. The belief that $p$ should be used as an unconditional premise in the agent's further reasoning and should guide what the agent does, which is what distinguishes a belief from a working hypothesis, whose consequences an agent draws only conditionally, in order to decide what to check next. The adverse judgment, for its part, should govern how the agent reports on and qualifies its conclusion, whether it requests verification, and what it says when asked whether the conclusion is established.

The dissociated configuration gives also a distinctive pattern of dependence that can be tested by intervening on the agent.

\begin{ex}[Double dissociation]\label{ex:double}
(a) Holding fixed the evaluator's warning and its record, manipulate the agent's signal: the agent's keeping of its conclusion changes, while its self-assessment does not. (b) Holding fixed the signal, vary the evaluator's record: the agent's self-assessment changes, while its keeping of conclusions in the cases where the signal is strong does not.
\end{ex}

A pattern of the kind described in (\ref{ex:double}) (a propos long used in neuropsychology as evidence that two capacities rest on separate mechanism) would show that each of the two states is governed by its own grounds and persists while the other varies, which is the best evidence available from behavior that the agent holds two attitudes with separate grounds, as opposed to a single attitude that it expresses inconsistently. These interventions are, moreover, already available: one can amplify or suppress internal representations directly, as in the experiments of Lindsey (2025).\footnote{The reversals documented by Laban et al.\ (2023) and Sharma et al.\ (2024), in which models abandon answers when users push back, are not cases of epistemic akrasia, since in these cases the first-order belief is given up.}

Of note, the design argument of Sections 5 and 6 is independent of how the attribution comes out. Even if the conflicting states are better described as the outputs of two processes than as attitudes of a single agent, a system resolving conflicts between those outputs by the veto or by rewarding agreement faces the difficulty identified in Section 5. Comparative Authority applies to it unchanged. The attribution settles whether the conflict is epistemic akrasia in the strict sense, and so consequently whether the epistemology of Section 2 bears on the agent as such or only on the relation between its components.

\section{8. Experimental Testing}

In this section, I shall venture into some programmatic remarks on experimentally testing these matters, on the understanding that a lot of nuance will have to be work out more thoroughly in future work.

As shown in Section 7, the design argument does not depend on how the attribution of attitudes comes out, but it is worth asking what does it depend on then, since that can reveal where the argument is exposed to evidence. I submit that it depends on the two claims contained in the definition of dissociation in (\ref{ex:dissociation}): first, that an agent's process for keeping its conclusions can use a signal that separates sound from defective reasoning even after a credible warning has arrived, and second, that the agents' self-assessment can lack access to that signal. As Section 3 showed, the available evidence motivates both claims without establishing that they hold together in the circumstance the argument requires, and since the recommendations of Section 6 also make predictions, it remains to describe how the claims and the predictions could be tested.

The first test has to do with the signal among warned cases, for these are the cases on which the argument turns. It requires investigations whose soundness is known, including true conclusions reached by defective reasoning and false conclusions reached by sound reasoning from misleading inputs, so that a signal which tracks soundness can be told apart from one which tracks only correctness, and it requires warnings whose correctness, whose source's demonstrated reliability, and whose assertiveness vary independently of one another, so that the credibility of a warning isn't tacitly identified with its truth. The question is how much more often a candidate signal (e.g. the confidence that Kumaran et al.\ (2025) measure or the output of a classifier of the kind that Orgad et al.\ (2025) train) is present in sound than in defective reasoning among the warned cases that the self-assessment rejects, compared with the threshold set by the proportion of well-founded conclusions in that class and the weights of Section 4. Once the assessment has information of its own, success among warned cases as a whole need not show that resisting its negative verdict improves retention.\footnote{In Section 4, among the cases rejected by the assessment, the signal is present in $57/60$ of the well-founded cases and $44/880$ of the ill-founded ones. Its likelihood ratio is $19$, above the equal-weight threshold $880/60=44/3$, and retaining the disputed conclusions earns $57-44=13$ units.} The question must be answered on datasets withheld from the training of any classifier. The second test asks whether intervening on the signal changes which conclusions the agent keeps, and whether the double dissociation of (\ref{ex:double}) obtains.

The third test compares interventions on a fixed collection of investigations, a collection withheld from all training that contains both sound reasoning and reasoning whose defects can be located, some of it drawn from the agent's own earlier work. The interventions are a veto, training that rewards agreement between conclusions and self-assessments, and integration that preserves the self-assessment's own checks, and each is compared with a matched control that lacks any incentive for agreement. The argument predicts that the veto will lower the proportion of well-founded conclusions that the agent keeps after misleading warnings, that the effect of training for agreement will depend on which of the two processes the training changes, and can be diagnosed only by measuring how well the self-assessment separates sound from defective reasoning on the fixed collection, and that integration will reduce the frequency of conflict while preserving the accuracy of the agent's keeping of conclusions. The fourth test, finally, separates a change in the self-assessment from a change in what the agent merely reports, for a model may learn to report approval while retaining the ability to locate an error when asked in another way.

\section{9. Conclusion}

To take stock, we started from the idea that self-correction requires an agent to take its own judgments about its reasoning seriously, even though those judgments can be mistaken. Artificial agents make this difficulty particularly acute since the process that leads to a conclusion may be better informed than the process assessing it. Enforcing agreement between the two processes can make the agent epistemically \textit{worse off} rather than better. A self-assessment can be reliable in general and still be the worse judge on just the cases where it would overrule the agent, since those cases are selected by disagreement, and the conflict between the two processes cannot by itself say which of them has erred. Thus, the possibility of epistemic akrasia in such systems, interestingly, reveals a more general limit of self-correction. An agent capable of correcting its conclusions ought to also be capable of correcting its judgments about those conclusions, at the same time as it preserves the independence that makes such judgments useful to begin with.

\newpage

\section*{References}
\begin{flushleft}
\small\setlength{\parskip}{4pt plus 1pt minus 1pt}
Africa, David Demitri, and Hans Ethan Ting. 2025. ``Does Self-Evaluation Enable Wireheading in Language Models?'' arXiv:2511.23092. \url{https://arxiv.org/abs/2511.23092}.

Arpaly, Nomy. 2000. ``On Acting Rationally against One's Best Judgment.'' \textit{Ethics} 110 (3): 488-513. \url{https://doi.org/10.1086/233321}.

Azaria, Amos, and Tom Mitchell. 2023. ``The Internal State of an LLM Knows When It's Lying.'' In \textit{Findings of the Association for Computational Linguistics: EMNLP 2023}.

Baker, Bowen, Joost Huizinga, Leo Gao, Zehao Dou, Melody Y. Guan, Aleksander Madry, Wojciech Zaremba, Jakub Pachocki, and David Farhi. 2025. ``Monitoring Reasoning Models for Misbehavior and the Risks of Promoting Obfuscation.'' arXiv:2503.11926. \url{https://arxiv.org/abs/2503.11926}.

Binder, Felix J., James Chua, Tomek Korbak, Henry Sleight, John Hughes, Robert Long, Ethan Perez, Miles Turpin, and Owain Evans. 2025. ``Looking Inward: Language Models Can Learn About Themselves by Introspection.'' In \textit{International Conference on Learning Representations (ICLR 2025)}. arXiv:2410.13787.

Broome, John. 1999. ``Normative Requirements.'' \textit{Ratio} 12 (4): 398--419.

Burns, Collin, Haotian Ye, Dan Klein, and Jacob Steinhardt. 2023. ``Discovering Latent Knowledge in Language Models Without Supervision.'' In \textit{International Conference on Learning Representations (ICLR 2023)}.

Chen, Wei-Lin, Zhepei Wei, Xinyu Zhu, Shi Feng, and Yu Meng. 2025. ``Do LLM Evaluators Prefer Themselves for a Reason?'' arXiv:2504.03846. \url{https://arxiv.org/abs/2504.03846}.

Christensen, David. 2010. ``Higher-Order Evidence.'' \textit{Philosophy and Phenomenological Research} 81 (1): 185--215.

Christensen, David. 2011. ``Disagreement, Question-Begging and Epistemic Self-Criticism.'' \textit{Philosophers' Imprint} 11 (6): 1--22.

Christensen, David. 2016. ``Disagreement, Drugs, etc.: From Accuracy to Akrasia.'' \textit{Episteme} 13 (4): 397--422. \url{https://doi.org/10.1017/epi.2016.20}.

Christiano, Paul, Ajeya Cotra, and Mark Xu. 2021. ``Eliciting Latent Knowledge: How to Tell If Your Eyes Deceive You.'' Technical report, Alignment Research Center.

Coates, Allen. 2012. ``Rational Epistemic Akrasia.'' \textit{American Philosophical Quarterly} 49 (2): 113--124.

Dorst, Kevin. 2020. ``Evidence: A Guide for the Uncertain.'' \textit{Philosophy and Phenomenological Research} 100 (3): 586--632.

Elga, Adam. 2007. ``Reflection and Disagreement.'' \textit{No\^us} 41 (3): 478--502.

Elga, Adam, and Agust\'in Rayo. 2021. ``Fragmentation and Information Access.'' In \textit{The Fragmented Mind}, edited by Cristina Borgoni, Dirk Kindermann, and Andrea Onofri, 37--53. Oxford: Oxford University Press. \url{https://doi.org/10.1093/oso/9780198850670.003.0002}.

Farquhar, Sebastian, Vikrant Varma, Zachary Kenton, Johannes Gasteiger, Vladimir Mikulik, and Rohin Shah. 2023. ``Challenges with Unsupervised LLM Knowledge Discovery.'' arXiv:2312.10029. \url{https://arxiv.org/abs/2312.10029}.

Feldman, Richard. 2005. ``Respecting the Evidence.'' \textit{Philosophical Perspectives} 19: 95--119.

Field, Claire. 2024. ``The Value of Incoherence.'' \textit{Philosophical Issues} 34 (1): 37--58. \url{https://doi.org/10.1111/phis.12266}.

Field, Claire. 2026. ``Embracing Incoherence.'' In \textit{Essays on Epistemic Dilemmas}, edited by Nick Hughes, 260--284. Oxford: Oxford University Press. \url{https://doi.org/10.1093/9780192585325.003.0010}.

Good, I. J. 1967. ``On the Principle of Total Evidence.'' \textit{The British Journal for the Philosophy of Science} 17 (4): 319--321. \url{https://doi.org/10.1093/bjps/17.4.319}.

Greco, Daniel. 2014. ``A Puzzle about Epistemic Akrasia.'' \textit{Philosophical Studies} 167 (2): 201--219.

Herrmann, Daniel A., and Benjamin A. Levinstein. 2025. ``Standards for Belief Representations in LLMs.'' \textit{Minds and Machines}. \url{https://doi.org/10.1007/s11023-024-09709-6}.

Horowitz, Sophie. 2014. ``Epistemic Akrasia.'' \textit{No\^us} 48 (4): 718--744. \url{https://doi.org/10.1111/nous.12026}.

Huang, Jie, Xinyun Chen, Swaroop Mishra, Huaixiu Steven Zheng, Adams Wei Yu, Xinying Song, and Denny Zhou. 2024. ``Large Language Models Cannot Self-Correct Reasoning Yet.'' In \textit{International Conference on Learning Representations (ICLR 2024)}.

Jackson, Elizabeth, and Peter Tan. 2022. ``Epistemic Akrasia and Belief--Credence Dualism.'' \textit{Philosophy and Phenomenological Research} 104 (3): 717--727. \url{https://doi.org/10.1111/phpr.12810}.

Kadavath, Saurav, et al. 2022. ``Language Models (Mostly) Know What They Know.'' arXiv:2207.05221. \url{https://arxiv.org/abs/2207.05221}.

Kelly, Thomas. 2010. ``Peer Disagreement and Higher-Order Evidence.'' In \textit{Disagreement}, edited by Richard Feldman and Ted A. Warfield, 111--174. Oxford: Oxford University Press.

Kolodny, Niko. 2005. ``Why Be Rational?'' \textit{Mind} 114 (455): 509--563.

Kumaran, Dharshan, Stephen M. Fleming, Larisa Markeeva, Joe Heyward, Andrea Banino, Mrinal Mathur, Razvan Pascanu, Simon Osindero, Benedetto De Martino, Petar Veli\v{c}kovi\'c, and Viorica Patraucean. 2025. ``How Overconfidence in Initial Choices and Underconfidence Under Criticism Modulate Change of Mind in Large Language Models.'' arXiv:2507.03120. \url{https://arxiv.org/abs/2507.03120}.

Kumaran, Dharshan, Nathaniel Daw, Simon Osindero, Petar Veli\v{c}kovi\'c, and Viorica Patraucean. 2026. ``Causal Evidence that Language Models Use Confidence to Drive Behavior.'' arXiv:2603.22161. \url{https://arxiv.org/abs/2603.22161}.

Laban, Philippe, Lidiya Murakhovs'ka, Caiming Xiong, and Chien-Sheng Wu. 2023. ``Are You Sure? Challenging LLMs Leads to Performance Drops in the FlipFlop Experiment.'' arXiv:2311.08596. \url{https://arxiv.org/abs/2311.08596}.

Lasonen, Maria. 2026. \textit{The Good, the Bad, and the Feasible: Knowledge and Reasonable Belief}. Oxford: Oxford University Press. \url{https://doi.org/10.1093/9780191881008.001.0001}.

Lasonen-Aarnio, Maria. 2010. ``Unreasonable Knowledge.'' \textit{Philosophical Perspectives} 24: 1--21.

Lasonen-Aarnio, Maria. 2014. ``Higher-Order Evidence and the Limits of Defeat.'' \textit{Philosophy and Phenomenological Research} 88 (2): 314--345. \url{https://doi.org/10.1111/phpr.12090}.

Lasonen-Aarnio, Maria. 2020. ``Enkrasia or Evidentialism? Learning to Love Mismatch.'' \textit{Philosophical Studies} 177 (3): 597--632.

Lasonen-Aarnio, Maria. 2021a. ``Dispositional Evaluations and Defeat.'' In \textit{Reasons, Justification, and Defeat}, edited by Jessica Brown and Mona Simion, 91--115. Oxford: Oxford University Press.

Lasonen-Aarnio, Maria. 2021b. ``Coherence as Competence.'' \textit{Episteme} 18 (3): 453--476. \url{https://doi.org/10.1017/epi.2021.33}.

Lindsey, Jack. 2025. ``Emergent Introspective Awareness in Large Language Models.'' \textit{Transformer Circuits Thread}. \url{https://transformer-circuits.pub/2025/introspection/index.html}.

Marks, Samuel, and Max Tegmark. 2024. ``The Geometry of Truth: Emergent Linear Structure in Large Language Model Representations of True/False Datasets.'' In \textit{Conference on Language Modeling (COLM 2024)}.

Orgad, Hadas, Michael Toker, Zorik Gekhman, Roi Reichart, Idan Szpektor, Hadas Kotek, and Yonatan Belinkov. 2025. ``LLMs Know More Than They Show: On the Intrinsic Representation of LLM Hallucinations.'' In \textit{International Conference on Learning Representations (ICLR 2025)}. arXiv:2410.02707.

Panickssery, Arjun, Samuel R. Bowman, and Shi Feng. 2024. ``LLM Evaluators Recognize and Favor Their Own Generations.'' In \textit{Advances in Neural Information Processing Systems 37 (NeurIPS 2024)}.

Schoenfield, Miriam. 2018. ``An Accuracy Based Approach to Higher Order Evidence.'' \textit{Philosophy and Phenomenological Research} 96 (3): 690--715. \url{https://doi.org/10.1111/phpr.12329}.

Shafayat, Sheikh, Fahim Tajwar, Ruslan Salakhutdinov, Jeff Schneider, and Andrea Zanette. 2025. ``Can Large Reasoning Models Self-Train?'' arXiv:2505.21444. \url{https://arxiv.org/abs/2505.21444}.

Sharma, Mrinank, et al. 2024. ``Towards Understanding Sycophancy in Language Models.'' In \textit{International Conference on Learning Representations (ICLR 2024)}.

Shea, Nicholas, Annika Boldt, Dan Bang, Nick Yeung, Cecilia Heyes, and Chris D. Frith. 2014. ``Supra-Personal Cognitive Control and Metacognition.'' \textit{Trends in Cognitive Sciences} 18 (4): 186--193. \url{https://doi.org/10.1016/j.tics.2014.01.006}.

Sliwa, Paulina, and Sophie Horowitz. 2015. ``Respecting All the Evidence.'' \textit{Philosophical Studies} 172 (11): 2835--2858.

Smithies, Declan. 2012. ``Moore's Paradox and the Accessibility of Justification.'' \textit{Philosophy and Phenomenological Research} 85 (2): 273--300.

Titelbaum, Michael G. 2015. ``Rationality's Fixed Point (or: In Defense of Right Reason).'' \textit{Oxford Studies in Epistemology} 5: 253--294.

Wang, Jiawei, Yanfei Zhou, Siddartha Devic, and Deqing Fu. 2026. ``Are LLM Decisions Faithful to Verbal Confidence?'' arXiv:2601.07767. \url{https://arxiv.org/abs/2601.07767}.

Weatherson, Brian. 2019. \textit{Normative Externalism}. Oxford: Oxford University Press.

Williamson, Timothy. 2014. ``Very Improbable Knowing.'' \textit{Erkenntnis} 79 (5): 971--999.

Worsnip, Alex. 2018. ``The Conflict of Evidence and Coherence.'' \textit{Philosophy and Phenomenological Research} 96 (1): 3--44. \url{https://doi.org/10.1111/phpr.12246}.

Worsnip, Alex. 2026. ``Evidence-Coherence Conflicts Revisited.'' In \textit{Essays on Epistemic Dilemmas}, edited by Nick Hughes, 230--259. Oxford: Oxford University Press. \url{https://doi.org/10.1093/9780192585325.003.0009}.

Xiong, Miao, et al. 2024. ``Can LLMs Express Their Uncertainty? An Empirical Evaluation of Confidence Elicitation in LLMs.'' In \textit{International Conference on Learning Representations (ICLR 2024)}.
\end{flushleft}
\end{document}